\documentclass{IOS-Book-Article}

\usepackage{mathptmx}
\usepackage[utf8]{inputenc}
\usepackage[colorinlistoftodos]{todonotes}
\usepackage{multirow}
\usepackage{float}
\usepackage{algorithm}

\usepackage{hyperref}
\usepackage{enumitem}
\usepackage{caption}

\def\hb{\hbox to 10.7 cm{}}

\begin{document}

\pagestyle{headings}
\def\thepage{}

\begin{frontmatter}              

\title{Argument Harvesting Using Chatbots}

\markboth{}{April 2018\hb}

\author[A]{\fnms{Lisa A.} \snm{Chalaguine}%
}
\author[B]{\fnms{Fiona L.} \snm{Hamilton}}
\author[A]{\fnms{Anthony} \snm{Hunter}}
\author[C]{\fnms{Henry W. W.} \snm{Potts}}
\runningauthor{Lisa Chalaguine}
\address[A]{Department of Computer Science, University College London, London, UK}
\address[B]{eHealth Unit, University College London, London, UK}
\address[C]{Institute of Health Informatics, University College
London, London UK}

\begin{abstract}
Much research in computational argumentation assumes that arguments and counterarguments can be obtained in some way. Yet, to improve and apply models of argument, we need methods for acquiring them. Current approaches include argument mining from text, hand coding of arguments by researchers, or generating arguments from knowledge bases. In this paper, we propose a new approach, which we call \emph{argument harvesting}, that uses a chatbot to enter into a dialogue with a participant to get arguments and counterarguments from him or her. Because it is automated, the chatbot can be used repeatedly in many dialogues, and thereby it can generate a large corpus. We describe the architecture of the chatbot, provide methods for managing a corpus of arguments and counterarguments, and an evaluation of our approach in a case study concerning attitudes of women to participation in sport. 
\end{abstract}
\end{frontmatter}
\markboth{April 2018\hb}{April 2018\hb}
\thispagestyle{empty}
\pagestyle{empty}

\section{Introduction}

Abstract argument graphs, such as proposed by Dung \cite{dung}, are an important formalism in computational models of argument. However, the issue of acquiring the graphs tends to be omitted. In order to construct graphs using \emph{real} arguments as opposed to theoretical, made-up scenarios, arguments have to be acquired from real-life sources. A common approach to argument acquisition assumes a static resource available on the internet where the topic of interest is/was already discussed. This, however, raises several problems: firstly, what if no discussion platform for a particular topic exists? Secondly, even if it exists, what if not enough representative people contribute to the opinion exchange? Thirdly, such platforms do not take into account attributes of the individuals who posited the arguments. The only solution is then to use questionnaires or to interview people directly. That, however, may be a labour-intensive and expensive undertaking. To address these issues, we believe that it is possible to automate the process of argument acquisition using chatbots. A chatbot is a computer program that can chat with humans via text. As a proof of concept, in this paper, we present a method focused on argument acquisition for behaviour change applications but which could be adopted to other application domains as well.

Recently chatbots have been developed for domains like health care and behaviour change \cite{healthcarechatbots2, healthcarechatbots1}. Human agents respond and converse with artificial agents in ways that to some extent mirror emotional and social discourse dynamics when discussing behavioural health \cite{fitz1}. Therefore, there is literature to suggest that using a chatbot to acquire user arguments on a certain behaviour and address the problems of traditional argument acquisition, is possible. However, existing chatbots that talk to users about their health and behaviour require domain-specific knowledge and usually focus on one topic such as healthy diet, exercising routine, stress management, and are not applicable across different domains. They are also not developed to acquire arguments. Nevertheless, as already shown by Weizenbaum, a chatbot that uses only generic questions is indeed capable of encouraging the user to talk about himself \cite{eliza}. His chatbot \emph{Eliza} simulated conversation by using pattern matching and pronoun substitution that gave users an illusion of understanding even though it had no built-in knowledge. This therefore indicates that generic questions may enable a chatbot to harvest arguments in diverse domains.   

So far, no attempts of using a chatbot for argument acquisition have been made in the computational argumentation domain. In this paper, we investigate the approach at \emph{argument harvesting} which we define as acquiring arguments with the help of a chatbot. We further perform three experiments with crowd-sourced participants in order to analyse the meaningfulness, values and relationships of the arguments. The contribution of our work is threefold: first, we describe a model for argument harvesting using a simple chatbot with little or no domain knowledge. Second, we show that people who give the same or a semantically similar argument, are most likely motivated by the same value when positing it. We demonstrate that it is therefore possible to train a classifier to predict the value (motivation) of an argument. And third, we present a method to cluster the harvested arguments by value and semantic similarity in order to automatically create several possibilities to counter a given argument. 

The rest of the paper is structured as follows: Section 2 gives some background theory on value-based argumentation frameworks, an overview on values, as well as our own definitions of values; Section 3 gives the aim of the paper and the hypotheses; Section 4 describes the chatbot architecture that was used for argument harvesting; Section 5 describes the experiments that were conducted throughout the study including their methodology and results, and in Section 6 we discuss and conclude our findings.


\section{Values in Argumentation}
In order to account for different points of view in debates, it has been recognised that the parties within a debate will have different perspectives on what is important to pursue, according to their subjective aspirations and preferences \cite{atkinson1}. In value-based argumentation \cite{VAF, VAF2} arguments promote specific values which account for the social interests of debate participants. Values are assigned to an argument when constructing argument graphs. They provide an explanation as to why it is not always possible to persuade others to accept an opinion simply by demonstrating facts and proofs. It may be that a particular individual will accept the facts of a decision but will reject the conclusion to act upon it because it does not support the values he or she holds \cite{atkinson1}.  Although we do not use value-based argumentation frameworks in this paper, we are interested in the notion of values and their relevance to argumentation in behaviour change.

None of the papers that apply value-based argumentation frameworks to specific examples \cite{atkinson2, nawabBC, mogdil1} explain where the values come from or according to what principles they should be chosen. We therefore need to define our own notion of values for our purpose. In a dialogue when someone posits an argument, they normally have some \emph{motivation} for choosing \emph{this} specific argument as opposed to another in that part of the dialogue. We call these \emph{categories of motivation} which are categories that are important to the \emph{life} of the agent. Note, we are not interested in motivations concerning the dialogue, e.g. winning the dialogue, revenge, showing-off, deceiving etc.

In this paper we study attitudes of women towards engaging in physical exercise. We are concerned with the notions of \emph{value} of an argument and the \emph{suitability} of a counterargument. The following example illustrates the two notions: given the categories of motivation for not exercising \emph{V = \{family, comfort, dignity, wealth\}}, suppose a woman (the persuader) is trying to convince her friend (the persuadee) to do more sports and gives the following argument:
\textbf{A1:}\emph{ “Physical activity is healthy and you should therefore go to the gym more often.”} The persuadee, assuming she is rational, will not try to counter the fact that physical activity is healthy and will most likely \emph{accept} that fact. She may, however, counter the conclusion (which action to take) with an argument that reflects her motivations (values) for not engaging in physical activity. She might say: \textbf{A2:} \emph{“I have no time because I have to look after my kids.”} In this case the argument promotes the value \emph{family}. To generalise this idea, we give the following definition for values which delineates how we can assign a value to an argument. 
\\
\\
\textbf{Definition 1} A \textbf{value assignment} by an agent to an argument \emph{A} is a category of motivation for the agent if the agent were to posit \emph{A}.
\\
\\
In the above definition we use the phrase ``if the agent were to posit \emph{A}" because we will investigate how individuals value arguments independently of a specific dialogue. 

In this paper, we are interested in a certain \emph{kind} of counterargument that is appropriate for dialogues in behaviour change. For such dialogues, we believe a counterargument should have the same value assignment as the argument it attacks. Continuing with the example above, the persuader would respect the value \emph{family} and give a counterargument A3 that attacks A2 but respects the value \emph{family}. For example: \textbf{A3:} \emph{“You could incorporate your children into your exercise routine. Like going roller blading in the park or swimming.”} So A3 attacks A2 while respecting the same value and still pursues the initial intention of persuading the persuadee to do more sports. This does not mean that the persuadee has to agree with the given counterargument, it merely means that the counterargument can be given as a \emph{suitable} counterargument to the previously posited argument. We define the notion of \emph{suitability} of a counterargument next:
\\
\\
\textbf{Definition 2} Let \emph{A} be an argument and let \emph{CA} be a counterargument that attacks \emph{A}. \emph{CA} is a \textbf{suitable} counterargument to \emph{A} iff \emph{A} has a value assignment \emph{V} and \emph{CA} has a value assignment \emph{V’} such that \emph{V = V’}.
\\
\\
In this section we have outlined the importance of values in argumentation for behaviour change and have given our own definition of \emph{value assignment} to arguments. We have also introduced the concept of \emph{suitability} for counterarguments that can be used to counter a previously posited argument that promotes a specific value. Given these notions of value and suitability, we want to test several hypotheses, given in the next section.

\section{Hypotheses}
In this paper, we make a first step towards argument harvesting. We chose \emph{attitudes of women to participation in sport} as a case study. We have developed a chatbot that harvests arguments and their values from women on why they do not engage in (more) physical activity. The chatbot also asks them to provide suitable counterarguments to their given arguments (more on the dialogue protocol in the next section). Each argument therefore has a value and a counterargument. Given this, we want to test three issues: first, whether different people are motivated by the same value if giving the same, or semantically similar argument. Second, whether our chatbot is capable of harvesting meaningful arguments i.e. those considered to be appropriate arguments by sufficiently many participants from the people group the argument was harvested from. Third, whether we can automatically match an argument with more suitable counterargument and therefore create more possibilities to counter a certain argument. We summarise these points in the following three hypotheses:

\begin{enumerate}[label=\textbf{H{\arabic*}}]

\item Different people that exposed to, but not necessarily posit, the same argument, assign to it the same or similar value, therefore making it possible to predict the value of an argument.

\item A domain neutral chatbot, with little or no domain specific knowledge, and by giving general responses, can acquire arguments that are perceived as meaningful by the people group the arguments were harvested from.

\item Given arguments semantically similar in meaning with the same value, counterarguments are interchangeable making it possible to use the counterargument of one argument as a counterargument to another argument.
\end{enumerate}

In the remainder of this paper we describe the design of our chatbot that was used for argument harvesting and explain the experiments conducted with the harvested arguments in order to test our hypotheses.

\section{Chatbot Design}

Messaging has become the most widely used communication layer on mobile platforms during the last few years, with Facebook Messenger (FM) being the most popular messaging application\footnote{1.3 billion active users as of December 2017}. FM is a free instant messaging service and software application which lets Facebook users chat with other users (or chatbots) on the main website as well as the mobile app. For building chatbots, the Messenger Send API gives the ability to send and receive messages. Due to the popularity of FM and the free API that Facebook provides we decided to use FM as the platform to deploy our chatbot.  

We created an application called \emph{ArgHealthBot} which users can send messages to. The application is linked to a Facebook page which has a \emph{Send Message} button\footnote{\url{https://www.facebook.com/Arghealthbot/}}. The page also displays a link to a website that contains the terms and conditions of the chatbot and states that we received ethical approval for our study, and a short description of the current experiment. For the screenshots of the website and the application, see Appendices G and H \cite{appendix}. When users click on \emph{Send Message}, a FM window pops up and allows them to send private messages to the application to which the chatbot is connected. The chatbot code is written in the Python programming language and consists of a Flask server and the text-processing code. The server code communicates with the Send API and the text-processing code processes the incoming messages from users and sends appropriate responses.

The dialogue protocol is the following: after the participant initiates the chat and consents to continue with the experiment, the chatbot asks to provide a reason for why she is not engaging in (more) physical activity, to which the participant answers with an argument (A1). If the chatbot considers the answer too short (less than 12 words) it asks to expand on the given argument. The chatbot queries the participant to expand on the argument only once. The pseudo-code and a description of the algorithm for query-generation (asking to expand on a given answer) can be found in Appendix I \cite{appendix}.

In order to assign values to the arguments we needed a set of values to choose from. We used the list of personal values from Scott Jeffrey\footnote{\url{https://scottjeffrey.com/core-values-list/}} as reference and pragmatically chose values that we found suitable. The values were: \emph{responsibility, comfort, dignity, satisfaction, relaxation, family, friendship, professionalism, productivity, wealth, knowledge, fun, recreation, ambition} and \emph{safety}. The chatbot presents the user with the list of values after she provided an argument and asks to choose the one she most associates with her argument. 

The chatbot then asks what the user would recommend a friend with the same problem. This is the counterargument to the previously given argument (CA1). The chatbot then picks up on that and asks why the user is not following her own advice. The user answers with another argument (A2). The chatbot asks again what she would advise a friend with the same problem (CA2). After collecting two argument-counterargument pairs the chatbot asks the participant whether she wants to continue or end the chat. Our chatbot therefore collects a minimum of two argument-counterargument pairs \{(A1, CA1), (A2, CA2)\}.


\section{Experiments}
In this section, we describe how we collected the arguments concerning women's participation in sports via argument harvesting (AH) and the experiments conducted with the harvested arguments. For each experiment we give the purpose, the methods used, the results and conclusion of our findings. The participants for all experiments were recruited via \emph{Prolific}, which is an online recruiting platform for scientific research studies. For each experiment we recruited from three disjoint groups: students (aged 18-25 and no children), women with children (aged 18-40 and not students) and women without children (aged 18-40 and not students), in the following referred to as the \emph{student, kids} and \emph{nokids} groups respectively. We opted for this division in order to get a wider spectrum of different arguments from different people groups, or \emph{audiences}. For each experiment we evaluate how the arguments are perceived by the audience it is meant for, based on the assumption that a particular argument is addressed to a specific audience \cite{audience}. The general prerequisites for taking part in our study were to be female, over 18 and engaged in less than 150 minutes of physical exercise per week. For the argument harvesting, we required the participants to have a Facebook account in order to chat with the chatbot. For the experiments, Google Forms were used.

\subsection{Argument Harvesting}

We conducted two rounds of argument harvesting (referred to as AH1 and AH2). In AH1, we used our chatbot to harvest arguments and their associated values and counterarguments from the three participant groups. 
In AH2, we harvested arguments and counterarguments without their values.

For AH1, we recruited 30 participants for the \emph{student} group, 30 for the \emph{kids} group and 50 for the \emph{nokids} group. The women who participated in the study and agreed to chat with the chatbot, initiated the conversations and the chat followed the dialogue protocol described in the previous section. For an example of a chat between participant and chatbot see Appendix E.1 \cite{appendix}.

Dialogues where participants described certain medical conditions like social anxiety, depression and scoliosis were removed from the data (10 dialogues in total). We decided that those require professional consultation and should not be included in this study. We also narrowed down the set of values by disregarding values that appeared in the whole data less than 5 times. The dialogues where at least one of the arguments had a deleted value, were removed (18 dialogues in total). The values used for the following experiments were: \emph{responsibility, family, productivity, dignity, wealth, comfort, relaxation} and \emph{fun}. 

For AH2, 20 participants from each group were recruited and asked to chat with the chatbot. This time we included more prerequisites during the recruitment, namely no chronic diseases, no long-term health conditions/disabilities and no ongoing mental illnesses. In this round, the chatbot did not ask the participants to assign values to their arguments. For an example of a chat, see Appendix E.2 \cite{appendix}. We harvested 40 arguments for each participant group in AH2 (no dialogues were deleted). The total number of argument-counterargument pairs after the two rounds of argument harvesting was 284 and can be found in Appendix A \cite{appendix}. 

After AH1, we made the following three observations. Firstly, some values were chosen more often than others and a smaller set of values therefore suffices to cover most of the arguments. Secondly, our simple chatbot was capable of harvesting a significant number of arguments. And lastly, we observed that many participants gave similar arguments or even the same argument, using different words. This opens the possibility of grouping arguments using clustering techniques. The experiments we conducted with the harvested arguments in order to test our hypotheses, are described in the following.

\subsection{Experiment I: Argument-Value Labeling}

The purpose of the experiment was to test whether different people assign the same (or similar) values to the same arguments that they have not posited themselves and whether it is possible to \emph{predict} the values of arguments by training a classifier and therefore verify Hypothesis I.
\begin{table}
\centering
\label{my-label}
\begin{tabular}{llllllrr}
\multicolumn{4}{l}{\begin{tabular}[c]{@{}l@{}}\textbf{Table 1.} Average agreement for values (V) and \\ parent-values (PV) for arguments harvested in \\ AH2.\end{tabular}} &                       & \multicolumn{3}{l}{\begin{tabular}[c]{@{}l@{}}\textbf{Table 2.} Accuracy of values (V) and \\parent-values (PV) when compared \\ to classifier-predicted values.\end{tabular}} \\ \cline{1-4} \cline{6-8} 
\multicolumn{1}{|l|}{Group}                 & \multicolumn{1}{l|}{S}             & \multicolumn{1}{l|}{K}            & \multicolumn{1}{l|}{NK}           & \multicolumn{1}{l|}{} & \multicolumn{1}{l|}{Group}                         & \multicolumn{1}{l|}{V Accuracy}                         & \multicolumn{1}{l|}{PV Accuracy}                         \\ \cline{1-4} \cline{6-8} 
\multicolumn{1}{|l|}{V Agreement}       & \multicolumn{1}{r|}{68.31\%}       & \multicolumn{1}{r|}{62.56\%}      & \multicolumn{1}{r|}{66.86\%}      & \multicolumn{1}{l|}{} & \multicolumn{1}{l|}{S}                             & \multicolumn{1}{r|}{50\%}                               & \multicolumn{1}{r|}{77.5\%}                              \\ \cline{1-4} \cline{6-8} 
\multicolumn{1}{|l|}{PV Agreement}          & \multicolumn{1}{r|}{81.45\%}       & \multicolumn{1}{r|}{86\%}         & \multicolumn{1}{r|}{81.43\%}      & \multicolumn{1}{l|}{} & \multicolumn{1}{l|}{K}                             & \multicolumn{1}{r|}{55\%}                               & \multicolumn{1}{r|}{82.5\%}                              \\ \cline{1-4} \cline{6-8} 
                                            &                                    &                                   &                                   & \multicolumn{1}{l|}{} & \multicolumn{1}{l|}{NK}                            & \multicolumn{1}{r|}{42.5\%}                             & \multicolumn{1}{r|}{70\%}                                \\ \cline{6-8} 
                                            &                                    &                                   &                                   & \multicolumn{1}{l|}{} & \multicolumn{1}{l|}{Avg}                           & \multicolumn{1}{r|}{49.2\%}                             & \multicolumn{1}{r|}{76.7\%}                              \\ \cline{6-8} 
\end{tabular}
\end{table}

The methods used in the experiment were the following. 20 participants for each group were recruited using the same prerequisites as for the argument harvesting, apart from the Facebook account, as no chatting with the chatbot was required. We used Google Forms for this task. Since we were interested in how the same group of people judged the arguments, we asked members of the \emph{student} group to assign values to the the arguments given by the students (respectively for the kids and nokids groups). The participants were presented the 40 arguments from their group harvested in AH2 and given a choice of 8 values. They were asked to ``read the argument for not engaging in physical activity and pick the value that they associated with the given argument”. The value that received the highest vote amongst the participant (value agreement) was chosen as the corresponding value for that particular argument. For example, if for argument A1, 16 out of 20 participants chose the value \emph{family}, then \emph{family} was assigned to A1 and the value agreement is 80\%. 

We observed that certain values are interchangeable: for example, the value `responsibility' was equivalent to `family' in the \emph{kids} group and `productivity' in the \emph{student} group. We therefore grouped six out of the eight values into the following two groups, calling these \emph{parent-values}: \textbf{CRF:} \emph{comfort}, \emph{relaxation} and \emph{fun}. \textbf{FRP:} \emph{family}, \emph{productivity} and \emph{responsibility}. The remaining two values \emph{wealth} and \emph{dignity} had no parent-value\footnote{They were grouped together as a parent-value during the classification in order to create a bigger group for the classifier as the two values on their own had too few examples.}. Parent-value agreement for the individual arguments was calculated by adding up the agreement rates for the individual values in that parent-value group. The agreement ratios for the individual groups (abbreviated \emph{S, K, NK} for the \emph{student, kids} and \emph{nokids} groups respectively) are shown in Table 1. On average, participants agreed 65.9\% on the value with an average standard deviation of 10.7 and 83\% agreed on the parent-value with a standard deviation of 11.2. For the individual arguments, see Appendix B \cite{appendix}. 

We used the values assigned by the participants that received the highest value agreement (participant values) to score the value-classifier. The arguments and values from AH1 were used for training, while the arguments from AH2 and the participant values were used for testing. We trained a Support Vector Machine with a linear kernel using the \texttt{scikit-learn} Python library\footnote{\url{http://scikit-learn.org/}}. We scored the classifier by comparing the classifier-predicted values to the participant values. The results are shown in Table 2. Accuracy is defined as the number of arguments where the value predicted by the classifier was the same as the participant value. There was a choice of 8 values and 3 parent-values. Random classification would therefore be 12.5\% and 33.33\% respectively. Our classifier had an average accuracy of 49.9\% for the values and 76.7\% for the parent-values. 

The accuracy of prediction for the \emph{nokids} group is lower than the other two groups due to the more diverse arguments compared to the \emph{kids} group. Table 3 shows how many arguments in each group are assigned with a specific parent-value. In the \emph{kids} group, 72.83\% of the arguments are assigned the values \emph{family} or \emph{responsibility}. These arguments often contain the words \emph{children, baby} and \emph{kids}. For the \emph{nokids} group the majority of the arguments (67.86\%) have the values \emph{comfort, relaxation} and \emph{fun}. Those arguments are much more diverse and do not have as many keywords in common which makes classification more difficult.
\begin{table}[H]
\centering
\label{my-label}
\caption{\textbf{Table 3.} Distribution of arguments (Args) with parent-values \emph{FRP, CRF}, and values \emph{Dignity} and \emph{Wealth} (the classifier-predicted values are used for the arguments from AH2).}
\begin{tabular}{|l|r|r|r|r|r|}
\hline
Group & \multicolumn{1}{l|}{No. of Args} & \multicolumn{1}{l|}{FRP} & \multicolumn{1}{l|}{CRF} & \multicolumn{1}{l|}{Dignity} & \multicolumn{1}{l|}{Wealth} \\ \hline
S     & 80                     & 31.25\%                  & 60\%                     & 1.25\%                       & 7.5\%                       \\ \hline
K     & 92                     & 72.83\%                  & 25\%                     & 0\%                          & 1.09\%                      \\ \hline
NK    & 112                    & 25.89\%                  & 67.86\%                  & 1.79\%                       & 3.57\%                      \\ \hline
\end{tabular}

\end{table}

It can be concluded that even though people might disagree on nuances like whether a certain argument promotes the value \emph{family} or \emph{responsibility} in the \emph{kids} dataset or cannot decide whether an argument given by a person is better associated with \emph{relaxation} or \emph{comfort}, the majority of people agree on the parent-value for a given argument. Out of 120 arguments only two arguments had a parent-value agreement below 60\% as can be seen in the tables in Appendix B \cite{appendix}. The results therefore support our Hypothesis I, that people independently assign the same or similar values to an argument that they have not posited themselves.

\subsection{Experiment II: Assessment of Harvested Arguments as Meaningful Arguments}
In this experiment, we wanted to assess whether a chatbot can be used as a tool for harvesting meaningful arguments and therefore verify Hypothesis II. 

The methods used in the experiment were the following. We recruited 10 participants for each group (like in the previous experiment, participants were representatives of the groups, e.g. students judging the arguments given by students). The prerequisites were the same as in Experiment I. Again no chatting with the chatbot was required and we used Google Forms for this task. Participants were presented all 40 arguments harvested of the corresponding group in AH2. We told the participants that the arguments were crowd-sourced reasons for not exercising and asked them whether they ``considered the given arguments as reasons they could give an appropriate advice". We also asked them to not judge the quality of the reason, rather just the completeness of it. After each argument they had the choice of selecting \emph{yes} or \emph{no}.

The results of the experiment are summarised in Table 4. We explain how we derived the results as follows: We set the threshold for considering a statement as an argument at 70\% annotator agreement (approval rate). This means that if a minimum of 7 out of the 10 participants answered the question whether a given statement is an argument positively, we labeled it as \emph{meaningful}. The student group had more arguments with an approval rate of 60\% than the other two groups, therefore closely missing our threshold and hence not being considered meaningful. For the results for the individual arguments, see Appendix C \cite{appendix}.

From the results, it can be concluded that a chatbot can indeed harvest meaningful arguments using no or very little domain knowledge, which supports our Hypothesis II. In total over 78\% of the arguments that were harvested in AH2 were considered meaningful. 

\begin{table}[H]
\centering
\caption{\textbf{Table 4.} Meaningful arguments (Args) in each group when the threshold is set to 70\% annotator agreement and above}
\label{my-label}
\begin{tabular}{|l|r|r|}
\hline
Group & \multicolumn{1}{l|}{No. of Args} & \multicolumn{1}{l|}{No. of meaningful Args}     \\ \hline
S & 40 & 28 (70\%)   \\ \hline
K  & 40  & 33 (82.5\%) \\ \hline
NK & 40  & 33 (82.5\%) \\ \hline
\end{tabular}
\end{table}

\subsection{Experiment III: Argument-Counterargument Matching}
The purpose of the experiment was to test Hypothesis III i.e. to evaluate whether the counterarguments of semantically similar arguments are interchangeable, making it possible to use the counterargument of one argument to counter another similar argument. 

The methods used in the experiment were the following. In order to cluster similar arguments we needed a clustering algorithm. Our dataset was too small to apply general-purpose unsupervised clustering algorithms, so we developed a specialised clustering algorithm that could take advantage of domain specific knowledge. We describe the algorithm below and the pseudo-code can be found in Appendix J \cite{appendix}.

First, we create a synonym list using WordNet \cite{wordnet}. This list contains lists of all the words in a given corpus that are synonyms of each other. Then the arguments are normalised by deleting stopwords and punctuation, and setting the case to low. We also delete exercise and time related words (\emph{exercise/s, sport/s, day/s, week/s, hour/s, thing/s, reason/s, main, lot}) because a lot of people repeated the chatbot's question in their answer (e.g.\emph{``The main reason I don't exercise is [...]"}). So we did not want to consider those in our similarity measurements. We also disregarded words that were used to describe how often they did or did not engage in a certain activity. Finally, for each argument, the noun phrases are extracted and stored as separate words and the synonyms are replaced with the first word in the corresponding synonym list. The arguments are stemmed in order to avoid treating different forms of a word as different words. After preprocessing the arguments, all arguments with the same value are clustered by comparing them to each other and clustering those together that share more than 50\% of the words. This results in clusters where each argument shares over 50\% of words with every other argument. An argument can occur in more than one cluster.

We applied the algorithm separately on the arguments of each participant group (see Appendix F for the clusters \cite{appendix}). Every argument has an original counterargument as given by the same participant during the chat with the chatbot. Each argument that appeared in a cluster (was `clustered') was matched with all the counterarguments from the other arguments in that cluster, apart from its original one. For example, if the arguments A1, A2, and A3 formed a cluster, then A1 would be matched with counterarguments of the other two arguments CA2 and CA3.

We evaluated the suitability of the counterarguments as follows: 10 participants for each group were recruited, with the same prerequisites as in Experiments I and II. We again used a Google Form where each argument was presented with its matched counterarguments and the participants were asked to choose which ones they believed was a suitable counterargument for the argument given. They were told that the arguments as well as the counterarguments were collected via crowd-sourcing and that they should not judge the quality of the arguments and counterarguments, but rather whether the counterargument is an appropriate response to the given argument.

\begin{table}[H]
\centering
\label{my-label}
\caption{\textbf{Table 5.} Total number of arguments (Args) in each group, number (percentage) of arguments clustered, the average number of counterarguments (CAs) per clustered argument and the number of argument clusters generated in each group.}
\begin{tabular}{|l|r|r|r|r|r|r|}
\hline
Group & \multicolumn{1}{l|}{No. of Args} & \multicolumn{1}{l|}{\begin{tabular}[c]{@{}l@{}}Clustered\\ total (\%)\end{tabular}} & \multicolumn{1}{l|}{\begin{tabular}[c]{@{}l@{}}Clustered\\ AH1 (\%)\end{tabular}} & \multicolumn{1}{l|}{\begin{tabular}[c]{@{}l@{}}Clustered\\ AH2 (\%)\end{tabular}} & \multicolumn{1}{l|}{\begin{tabular}[c]{@{}l@{}}Avg\\ CAs\end{tabular}} & \multicolumn{1}{l|}{Clusters} \\ \hline
S     & 80    & 40 (50\%)     & 18 (45\%)     & 22 (55\%)    & 3.65     & 19 \\ \hline
K     & 92  & 49 (53.26\%)    & 23 (44\%)  & 26 (65\%)    & 7.39   & 22  \\ \hline
NK    & 112      & 42 (37.5\%)    & 24 (33\%)     & 16 (40\%)      & 6.62                                                                  & 14    \\ \hline
\end{tabular} 

\end{table}

The results of the experiment are summarised in Tables 5-7. Table 5 shows how many arguments were clustered in the individual groups and the two rounds of harvesting. We can see that in the \emph{nokids} group fewer arguments were clustered than in the other two groups. This is due to the higher diversity in arguments and more complex synonyms.

The counterarguments of each argument received a certain approval rate, showing how often a given counterargument was selected by a participant. Table 6 (column 3) shows the average approval rates of the counterarguments for each argument in that group. For example if an argument had three counterarguments and the approval rates of them were 20\%, 70\% and 90\%, the average approval rate of the counterarguments for that argument would be 60\%. For more examples see Appendix D \cite{appendix}.

We considered the average number of suitable counterarguments per argument by using an approval rate threshold of 50\%. If, for instance, an argument had three counterarguments with the approval rates 40\%, 50\% and 60\% respectively, the second and third would be considered suitable and the number of suitable counterarguments would be 66.7\% (2/3). The results are shown in Table 6 (column 4). The reason for the lower threshold is the high variance of quality amongst counterarguments. Some counterarguments scored poorly because they give inappropriate advice (see Example 1).

\begin{table}[H]
\centering

\label{my-label}
\begin{tabular}{lrrrllrr}
\multicolumn{4}{l}{\begin{tabular}[c]{@{}l@{}}\textbf{Table 6.} Average approval rate (AR) of counterarguments (CAs)\\ per argument and the average number of suitable CAs per argu-\\ment with approval threshold of 50\%.\end{tabular}} &                       & \multicolumn{3}{l}{\begin{tabular}[c]{@{}l@{}}\textbf{Table 7.} The average approval\\ rate (AR) of individual counter-\\ arguments (CAs) when matched\\ with the corresponding argu- \\ ments in their cluster.\end{tabular}} \\ \cline{1-4} \cline{6-8} 
\multicolumn{1}{|l|}{Group}                         & \multicolumn{1}{l|}{No. of Args}                         & \multicolumn{1}{l|}{Avg. CA AR}                         & \multicolumn{1}{l|}{Avg. No. suitable CAs}                         & \multicolumn{1}{l|}{} & \multicolumn{1}{l|}{Group}                                        & \multicolumn{1}{l|}{No. of CAs}                                              & \multicolumn{1}{l|}{Avg. AR}                                       \\ \cline{1-4} \cline{6-8} 
\multicolumn{1}{|l|}{S}                             & \multicolumn{1}{r|}{40}                                  & \multicolumn{1}{r|}{70.37\%}                            & \multicolumn{1}{r|}{80.66\%}                                         & \multicolumn{1}{l|}{} & \multicolumn{1}{l|}{S}                                            & \multicolumn{1}{r|}{40}                                                      & \multicolumn{1}{r|}{69.18\%}                                       \\ \cline{1-4} \cline{6-8} 
\multicolumn{1}{|l|}{K}                             & \multicolumn{1}{r|}{49}                                  & \multicolumn{1}{r|}{69.04\%}                            & \multicolumn{1}{r|}{84.41\%}                                         & \multicolumn{1}{l|}{} & \multicolumn{1}{l|}{K}                                            & \multicolumn{1}{r|}{46\footnotemark}                                        & \multicolumn{1}{r|}{72.01\%}                                       \\ \cline{1-4} \cline{6-8} 
\multicolumn{1}{|l|}{NK}                            & \multicolumn{1}{r|}{42}                                  & \multicolumn{1}{r|}{60.10\%}                            & \multicolumn{1}{r|}{78.89\%}                                         & \multicolumn{1}{l|}{} & \multicolumn{1}{l|}{NK}                                           & \multicolumn{1}{r|}{42}                                                      & \multicolumn{1}{r|}{58.82\%}                                       \\ \cline{1-4} \cline{6-8} 
\end{tabular}
\end{table}

\footnotetext{There are only 46 counterarguments for the 49 clustered arguments because in three cases the participants answered \emph{``I don't know"} instead of giving a counterargument.}

We also analysed the approval rate that the individual counterarguments received, averaging all the approval rates that a counterargument received for all the arguments it was matched with. This way we wanted to identify inadequate  counterarguments and wrongly clustered arguments. For example, if counterargument CA4 was matched with three arguments A1, A2, A3 and received an approval rate of 40\% for A1, 50\% for A2 and 80\% for A3, the average approval rate for CA4 would be 56.7\%. The results are shown in Table 7. The following is an example of an inappropriate counterargument:
\\
\\
\textbf{Example 1} The argument A4 and counterargument CA4 were given by the same partici-pant. 

A4: \emph{``I only sometimes do sports because I am too busy and tired from my uni work".}

CA4: \emph{``You could join a sport team with a friend or find a gym buddy".}
\\
A4 was clustered with similar arguments (a total of 6) and therefore CA4 was matched with all the arguments of that cluster. It was, however, never approved as a suitable counterargument and had the lowest average approval rate in the \emph{student} dataset (17.5\%). It is not surprising that this counterargument was not considered a good one. It does not advise on how to manage your time better and/or emphasise the benefits of physical exercise. In the chat, when the chatbot asked why the person was not following her own advice, the participant indeed answered: \emph{``like I said, I am often too busy to do so. I mostly study or try to catch up on sleep"}. A counterargument that can be countered with \emph{``like I said..."} is unlikely to be an appropriate counterargument. 
\\

From the results in this section, it can be seen that counterarguments of similar arguments are interchangeable as long as they give appropriate advice, which supports our Hypothesis III about the interchangeability of counterarguments of semantically similar arguments. With the current data participants perceive a counterargument from a similar argument as suitable about 80\% of the time, when we set the threshold for suitability at 50\% approval rate. Regarding the clustering algorithm, only 131 out of the 284 arguments were clustered. This was due to several factors including wrong classification by the value-classifier, more complex synonyms including more than one word and lost negations in preprocessing of the arguments, specific explanations for a common reason, implicit meanings and specific arguments that did not repeat within the data. In the next section we discuss the results of our experiments.

\section{Discussion}
Our contribution in this paper is threefold. Firstly, we have shown that a simple chatbot with little or no domain knowledge can acquire meaningful arguments. We have focused on the behaviour change domain, 
where ordinary people give simple arguments that are nevertheless full of meaning and importance. They are the kind of arguments that have been neglected in the formal as well as informal argumentation literature. There is little literature on how to analyse this sort of argument and even less on how to acquire them. 

Secondly, we have shown that people mostly assign the same or a similar value to given arguments which makes it possible to predict values of arguments with the help of a classifier. Given this observation, it can be concluded that given an argument, most people will be motivated by the same value if positing it. We also made a first attempt of finding a suitable set of categories of motivation (values) for a specific topic, by letting the participants assign the values to their arguments themselves.

Thirdly, we presented a method to cluster arguments by values and similarity in order to create several possibilities to counter a given argument and evaluated whether the counterarguments of those are interchangeable. The results show that this is the case, given the counterargument itself is appropriate. However, in the future we want to research other methods of counterargument acquisition. One possibility is to harvest the arguments from one group (e.g. people who do not do sports) and acquire the counterargument from the group with the opposite behaviour (people who do sports). 


Argument harvesting can potentially be used in other domains such as politics, culture and marketing. The harvested arguments could then be used by politicians to evaluate a political decision, by theater directors to find out what people think about their play or by sales managers in order to get feedback on a new product.

Our future aim is to develop a chatbot for persuading people to change behaviour (e.g. to do more exercise) that can predict the value of an argument given by the user and answer with a suitable counterargument by querying a previously harvested corpus of arguments, counterarguments and their corresponding values. This chatbot would compare the argument to other arguments in the corpus, match it with similar ones, creating a cluster, and reply with one of the suitable counterarguments from that cluster. The choice of several counterarguments for an argument gives the chatbot the opportunity to backtrack and present another counterargument as soon as the chatbot gets `stuck' when the user gives an argument that cannot be matched to any argument in the corpus.

\section*{Acknowledgements}
We would like to thank Trevor Bench-Capon for his helpful insights on values in value-based argumentation. The first author is funded by a PhD studentship from the EPSRC.


\bibliography{biblio}
\bibliographystyle{abbrv}


\end{document}